\documentclass[conference]{IEEEtran}
\IEEEoverridecommandlockouts
\usepackage{cite}
\usepackage{amsmath,amssymb,amsfonts}
\usepackage{graphicx}
\usepackage[ruled,vlined,linesnumbered]{algorithm2e}
\usepackage{textcomp}
\usepackage{xcolor}
\usepackage[caption=false]{subfig}
\usepackage{enumitem}
\usepackage{balance}
\usepackage{float}
\usepackage{placeins}
\usepackage{etoolbox}

\AtBeginEnvironment{table}{%
  \begingroup
  \setlength{\intextsep}{0.5\dimexpr\intextsep\relax}%
}
\AfterEndEnvironment{table}{\endgroup}
\AtBeginEnvironment{table*}{%
  \begingroup
  \setlength{\dbltextfloatsep}{0.5\dimexpr\dbltextfloatsep\relax}%
}
\AfterEndEnvironment{table*}{\endgroup}
\usepackage{array}
\usepackage{tabularx}  
\usepackage{booktabs}  
\usepackage{arydshln} 
\usepackage{multirow}
\usepackage{makecell}
\renewcommand\arraystretch{1}

\def\BibTeX{{\rm B\kern-.05em{\sc i\kern-.025em b}\kern-.08em
    T\kern-.1667em\lower.7ex\hbox{E}\kern-.125emX}}
\begin{document}

\title{User Representation via Cross Multi-source Behavior Pre-training for Mobile Games}
\author{
  \IEEEauthorblockN{
    Chengqi Yang\IEEEauthorrefmark{2}\IEEEauthorrefmark{3},
    Yiran Qiao\IEEEauthorrefmark{2}\IEEEauthorrefmark{3},
    Feng Liu\IEEEauthorrefmark{4},
    Xingyu Lou\IEEEauthorrefmark{4},
    Zijun Zhou\IEEEauthorrefmark{4},
    Xiaoyun Mo\IEEEauthorrefmark{4}\\
    Changwang Zhang\IEEEauthorrefmark{4},
    Jiayuan Xu\IEEEauthorrefmark{2}\IEEEauthorrefmark{3},
    Jun Wang\IEEEauthorrefmark{4},
    Xiang Ao\IEEEauthorrefmark{2}\IEEEauthorrefmark{4}\IEEEauthorrefmark{5}\IEEEauthorrefmark{1}
    \thanks{\IEEEauthorrefmark{1}Xiang Ao is the corresponding author. (Email: aoxiang@ict.ac.cn)}
    \thanks{This work was done while Chengqi Yang was at OPPO Research Institute.}
  }

  \IEEEauthorblockA{
    \IEEEauthorrefmark{2} \textit{State Key Laboratory of AI Safety, Institute of Computing Technology, Chinese Academy of Sciences (CAS), Beijing, China.}
  }
  \IEEEauthorblockA{
    \IEEEauthorrefmark{3} \textit{University of Chinese Academy of Sciences, CAS, Beijing, China.}
  }
  \IEEEauthorblockA{
    \IEEEauthorrefmark{4} \textit{OPPO Research Institute, Shenzhen, China}
  }
  \IEEEauthorblockA{
    \IEEEauthorrefmark{5} \textit{Institute of Intelligent Computing Technology, CAS, Suzhou, China.}
  }
}

\maketitle

\begin{abstract}

User representation pre-training has become a fundamental paradigm for alleviating data sparsity in downstream personalization tasks. However, existing studies predominantly focus on single-app or app-level behaviors, overlooking the inherently cross-source and multi-granular nature of user activities on mobile devices. At the device level, user intent emerges from complex interactions among heterogeneous behavior sources and hierarchical action structures, posing challenges that cannot be addressed by conventional app-centric modeling. 
To tackle this issue, we propose \textbf{CM-PTM}, a novel \textbf{C}ross \textbf{M}ulti-source Behavior \textbf{P}re-\textbf{T}raining \textbf{M}odel tailored for mobile game user representation learning on device-level behavioral logs. CM-PTM employs hierarchical cascaded mask-then-predict proxy tasks that first infer the source of the next behavior and then progressively refine predictions at the app-action level. This design enables unified modeling of cross-source dependencies and fine-grained behavioral dynamics within a single pre-training paradigm. 
Extensive experiments on large-scale real-world mobile datasets demonstrate that CM-PTM effectively captures users' endogenous interests and consistently delivers significant performance gains on downstream mobile game recommendation tasks.

\end{abstract}


\section{Introduction}\label{sec:intro}

User representation modeling~\cite{perceive} aims at learning dense vector embeddings that capture user characteristics and interests from historical behaviors. 
The learned user representations can be applied to various downstream user-related tasks, such as user profile prediction~\cite{combination}, personalized recommendation~\cite{2019deep}, fraud detection~\cite{liu2022user}, and click-through-rate prediction~\cite{kalman}.

Inspired by the success of pre-training models~(PTMs) in NLP~\cite{simcse,r} and CV~\cite{exploring,unsupervised}, some studies have begun to leverage large-scale unlabeled behavioral data to pre-train user representations via self-supervised learning.
For example, several methods~\cite{peterrec,ptum} treat user behavior sequences as word sequences, adopting masked-behavior prediction and next-behavior prediction objectives. Some studies~\cite{clue,cl4rec,corec} employ contrastive learning by maximizing the consistency between original and augmented views of user behaviors. 

Despite the success of previous studies, we observe that they are predominantly confined to single-source or app-level behavior modeling~\cite{adaptssr,peterrec,cl4rec,clue}.
In this paper, we aim to learn user representations by leveraging all available behaviors from a user's mobile device. Real-world mobile devices naturally aggregate user behaviors from multiple heterogeneous sources with diverse granularity and accessibility constraints. 
Therefore, device-level user representation learning cannot be reduced to a simple extension of app-centric modeling, but instead constitutes a distinct problem setting with intrinsic structural complexity.

Specifically, under privacy and permission constraints, mobile devices typically observe two categories of behavioral data: (1) system-level applications~(e.g., App Store, Game Center, Browser) that are pre-installed with the OS, for which providers have privileged access to fine-grained interaction data; (2) third-party applications installed by users, for which providers can only observe coarse-grained events such as installation, launch, and uninstall, without access to in-app interactions. Notably, user interactions with third-party applications are far more frequent than those with system-level ones, leading to a pronounced mismatch between behavioral volume and behavioral granularity. This asymmetry introduces substantial challenges in learning high-quality user representations from mobile device data.

\begin{figure}[t]
  \centering
  \includegraphics[width=0.88\linewidth]{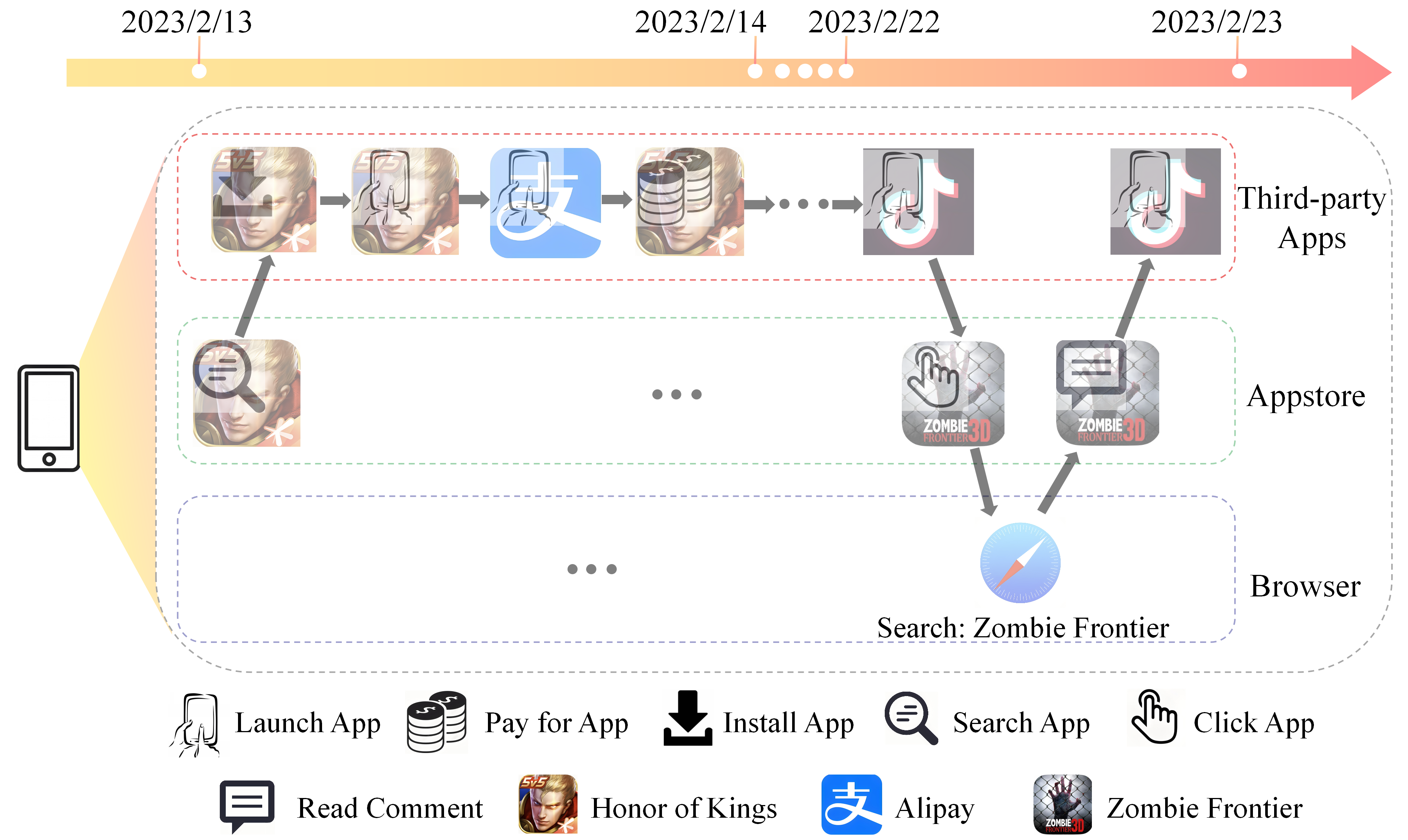}
  \caption{
  An illustrative example of user behavior on a mobile device from our real-world dataset. `Browser' indicates interactions within the pre-installed browser; `Appstore' denotes activities in the system App Store; `Third-party Apps' represent coarse-grained interactions regarding user-installed applications.}
  \label{fig:ex}
\end{figure}

More fundamentally, user intentions on mobile devices often emerge from cross-source interactions spanning heterogeneous behavior streams. 
Figure~\ref{fig:ex} illustrates a real user case from our dataset. The user's interactions with the games \emph{Honor of Kings} and \emph{Zombie Frontier} span different sources, including the third-party App, the App Store, and the Browser. A closer inspection reveals that the user may have encountered an advertisement for \emph{Zombie Frontier} on \emph{TikTok}, explored its details via the App Store and Browser, but ultimately did not download the game, suggesting weak installation intent despite multi-step engagement across sources.

This example underscores the potential and complexity of constructing comprehensive user representations by integrating behaviors from multiple, semantically diverse sources. Unlike prior studies that typically assume single-source behavior sequences, mobile device data entails cross-source interactions where user intentions are distributed across behaviors of differing types and granularities. Directly aggregating all behaviors into a single sequence~\cite{clue,ccl,ptum} fails to account for these structural distinctions, often degrading model performance~(as shown in the ablation study). Moreover, cross-source interleaving weakens local behavioral continuity within individual sources, undermining the core assumption of strong intra-sequence dependencies underlying many pre-training frameworks~\cite{peterrec,ptum,bert4rec}.

To address these issues, we propose a novel user representation pre-training framework tailored to mobile game recommendation on device-level multi-source logs, termed CM-PTM~(\underline{C}ross \underline{M}ulti-source Behavior \underline{P}re-\underline{T}raining \underline{M}odel). CM-PTM takes as input multi-source, multi-granularity user behavior sequences organized by their respective data sources. Motivated by a hierarchical decision structure for progressive intent refinement, CM-PTM employs a cascaded self-supervised prediction strategy. Given a user’s historical behaviors, it first predicts the source of the next behavior, then progressively infers the App category, the specific App ID, and the user's operation type within the App if applicable. Each stage is conditioned on the outcome of the previous one, enabling the model to incrementally refine its understanding of user intent across heterogeneous behavior streams.
Crucially, CM-PTM is designed as a system-level architecture for asymmetric cross-source fusion rather than as a collection of new attention primitives, enabling unified representation learning under the volume--granularity mismatch inherent to mobile device logs.
We conducted extensive experiments on large-scale real-world datasets in the game recommendation scenario, demonstrating that CM-PTM achieves substantial performance improvements.

Our contributions can be summarized as follows:

\begin{itemize}[leftmargin=*]
\item We investigate user representation pre-training based on behavioral data from mobile devices. In this scenario, user intentions may span multiple behavior sequences with different sources and granularities, which limits the direct applicability of conventional single-sequence pre-training paradigms to this setting.
\item We propose a pre-training approach called CM-PTM, which is built upon the mask-then-predict framework. CM-PTM incorporates several hierarchical cascaded proxy tasks specifically designed to capture cross-source associations, thereby improving the representation of user behavior across multiple sources.
\item Both extensive offline experiments and online A/B tests are conducted to demonstrate the effectiveness and robustness of our proposed approach in large-scale real-world industrial scenarios.
\end{itemize}
\section{Related Work} \label{sec:related}
In this section, we briefly review several existing works, including user representation pre-training, multi-behavior modeling and cross-domain recommendation.

\subsection{User Representation Pre-training}
The pre-training paradigm 
has been widely used in NLP~\cite{bert,xlnet,elmo} and CV~\cite{cv01,cv02}, with~\cite{roberta} attributing its effectiveness to the mask-then-predict strategy. 
Inspired by these successes, the pre-training paradigm has been used to learn user representations from large-scale unlabeled behaviors. Existing methods can be categorized into generative and discriminative methods.
Generative methods typically disrupt and reconstruct user behavior sequences. Commonly used strategies include mask behavior prediction~(MBP)~\cite{bert4sessrec,uprec,ubert,liu2022user,ptum} and next behavior prediction~(NBP)~\cite{peterrec,uaf,ptum}. 
~\cite{robust,zhang2024pre} replaced NBP with a behavior distribution prediction to reduce the negative influence of noise in user behaviors. 
Discriminative methods maximize the similarity between two behavioral views through contrastive learning. For example, ~\cite{zuo2024plip,lin2024multi} maximize the similarity between multi-modal or multi-granularity behavioral views, and other studies use data augmentation strategies to obtain different views.
Commonly used data augmentation strategies include explicit methods, such as masking~\cite{ccl,cl4rec}, cropping~\cite{crop}, and substitution~\cite{corec}, and implicit methods, where the same behavior sequence is fed to the model twice with different dropout masks~\cite{clue,duorec,simcse,luo2024perfedrec++}.
However, these pre-training methods cannot capture the cross-source correlations, making them less suitable for our scenario.

\subsection{Multi-Behavior Modeling}
Multi-behavior modeling aims to leverage various kinds of user behaviors to provide a more comprehensive understanding of user interests. \cite{atrank,online,wang2020beyond,jin2020multi,zhao2025joint} have demonstrated that multi-behavior modeling outperforms single-behavior approaches.~\cite{wyq,xu2023multi,zhang2023shgcn} use graph neural networks~(GNNs) to mine the complex relationships between heterogeneous user behaviors while~\cite{lin2024multi,elsayed2024multi} use Transformer-based architectures~\cite{trm} to model different behavior sequences, then capture the correlation between heterogeneous behaviors. Additionally,~\cite{shao2024filter} combined GNN-based and Transformer-based methods to model users' periodic behavior patterns and the dependencies between multiple behaviors. 
Although these approaches have achieved good results, they all focus on the behavior of a single source, potentially limiting their ability in our settings.

\subsection{Cross-Domain Recommendation}
The core concept of cross-domain recommendation~(CDR) is to integrate and analyze user behaviors and preferences in different domains to provide personalized recommendations.
Some CDR methods integrate user behaviors of different granularities from multiple domains to improve the accuracy of the recommendation system~\cite{zhang2025comprehensive}. Mainstream methods include RNN-based, GNN-based and attention-based methods.~\cite{chen2024novel} proposed an RNN-based long short-term memory fusion module that integrates knowledge extracted from multiple domains. GNN-based methods usually use graph convolution to aggregate neighborhood information~\cite{guo2021gcn} or graph attention networks~\cite{li2023preference} to fuse cross-domain semantic associations.~\cite{yang2024federated} introduces differential privacy protection for cross-domain knowledge transfer on this basis.
Attention-based methods usually optimize cross-domain feature interactions through hierarchical attention mechanisms~\cite{liang2022hierarchical}, domain adaptation~\cite{jiang2022adaptive}, and multiple contrastive learning~\cite{ma2024triple}.
However, existing methods focus on how to efficiently integrate multi-granularity behaviors from multiple domains without explicitly capturing the cross-source correlations of user behaviors.

\section{Problem Statement}
\label{sec:problem}

We assume that a substantial amount of behavioral data from $n$ different sources can be accessed on mobile devices. Let $\mathcal{U} = \{u_1, u_2, ..., u_{\vert \mathcal{U} \vert}\}$ denote the set of users, and $\mathcal{V}_k$, where $k=1,...,n$, denotes the set of all the behaviors from the $k$-th source, respectively. For each $u \in \mathcal{U}$, $\mathcal{S}_{u}^{k} = \{v_{<u,1>}^k, ..., v_{<u,t>}^k, ..., v_{<u,\vert \mathcal{S}_{u}^{k} \vert>}^k \}$ denotes their chronologically ordered sequence of behaviors from the $k$-th source.

\textbf{Notation.} Throughout the paper, italic $u \in \mathcal{U}$ denotes a user entity only. Bold $\boldsymbol{u}^k \in \mathbb{R}^{2d}$ denotes the source-specific user representation used in pre-training, and bold $\boldsymbol{u} \in \mathbb{R}^{d_u}$ denotes the concatenated downstream representation; neither bold symbol denotes the user ID. Subscripts such as $u$ in $\mathcal{S}_u^k$, $\boldsymbol{E}_u^k$, and $\mathcal{L}^{u}$ always refer to the user entity.

Our task aims to learn a function $\mathcal{F}_\theta$ that maps multi-granularity user behaviors from multiple sources into a dense user representation, formulated as
\begin{equation}
\label{eq:user}
    \boldsymbol{u} = \mathcal{F}_\theta(\mathcal{S}_u^1, ..., \mathcal{S}_u^n) \in \mathbb{R}^{d_u},
\end{equation}
where $\boldsymbol{u}$ denotes the learned downstream representation for user $u \in \mathcal{U}$, ${d_u}$ denotes its dimension, and $\theta$ denotes the model parameters.

\section{Methodology}
\label{sec:method}

\begin{figure*}[htbp]
\centering
\includegraphics[width=0.85\linewidth]{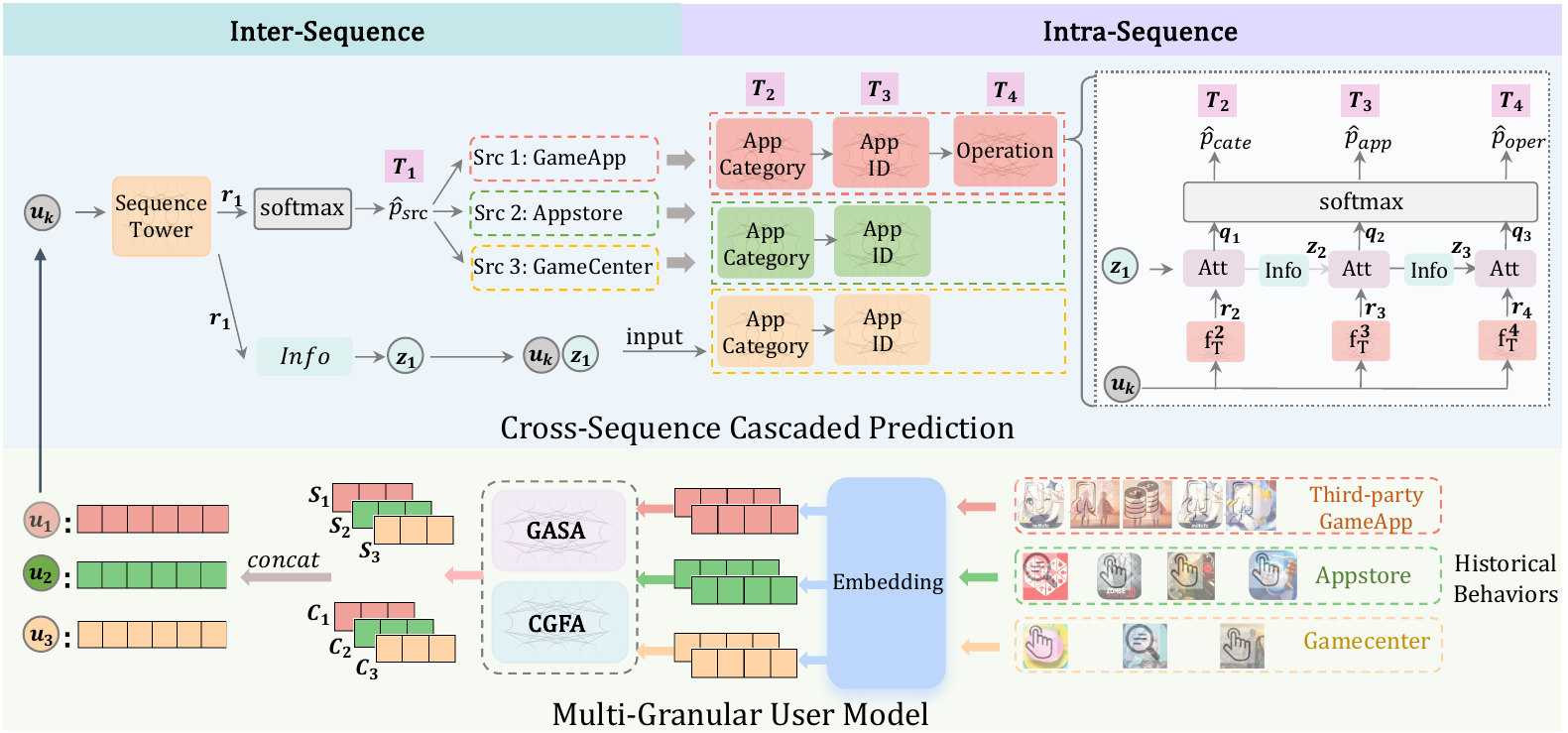}
  \caption{The overall architecture of the proposed CM-PTM framework. The lower part depicts the multi-granularity user encoder, which derives user representations from historical behaviors in different sources. The upper part shows the cross-source cascaded prediction module, consisting of inter-source and intra-source stages. Bold $\boldsymbol{u}^k$ in the figure denotes the source-specific representation in Eq.~(\ref{eq:user_repr}); italic $u$ denotes the user entity (Sec.~\ref{sec:problem}).}
\label{fig:model}
\end{figure*}

In this section, we describe CM-PTM in detail, as illustrated in Figure~\ref{fig:model}. It comprises two components: the multi-granularity user model and the cross-source cascaded prediction module.
The former integrates fine-grained and coarse-grained behaviors through hierarchical attention mechanisms, while the latter gradually refines the prediction of users' next behavior, from source prediction to fine-grained action prediction.

\subsection{Multi-Granularity User Model}
The key challenge lies in simultaneously preserving intra-source behavioral continuity and capturing inter-source intent correlations under heterogeneous granularity.
Unlike conventional multi-behavior modeling where all actions are observed at comparable granularity within a unified platform, device-level logs exhibit a severe mismatch between behavioral volume and granularity across system apps~(fine-grained) and third-party apps~(coarse-grained).
Our primary contribution is therefore not the invention of novel attention operators, but an architectural design tailored to this asymmetric industrial constraint.
We first use a behavior embedding module to transform behavior sequences into fixed-length embedding sequences. Then, to integrate multi-granularity behaviors from multiple sources under this asymmetry, we design a granularity-aware self-attention mechanism~(GASA) and a cross-granularity fusion attention mechanism~(CGFA), whose integration explicitly targets asymmetric cross-source fusion---a setting where standard Transformers that treat all behavior tokens uniformly fail to model effectively---and fuse their outputs as the user representation.

\subsubsection{Behavior Embedding}

In this paper, we collect user behaviors from OPPO device-level logs under three distinct sources: third-party GameApps, Appstore, and Gamecenter. Our method is instantiated on this fixed multi-source, multi-granularity setting in mobile game recommendation. 

For $u \in \mathcal{U}$, we obtain the behavior embeddings of user $u$ as:
\begin{equation}
\begin{split}
    \boldsymbol{E}_{u}^{k} &= \mathrm{Embedding}(\mathcal{S}_{u}^{k}) \\
    &=[\boldsymbol{e}_{<u,1>}^k,...,\boldsymbol{e}_{<u,t>}^k,...,\boldsymbol{e}_{<u,\vert \mathcal{S}_u^k \vert>}^k] \in \mathbb{R}^{\vert \mathcal{S}_u^k \vert \times d},
\end{split}
\end{equation}
where $d$ is the embedding dimension, $\mathcal{S}_u^k$ represents the behavior sequence of user $u$ from the $k$-th source, $[\cdot]$ denotes the concatenation of per-event embeddings, and $\boldsymbol{e}_{<u,t>}^k$ is the embedding of the $t$-th behavioral event in source $k$.

We set the maximum length of the embedding sequence $\boldsymbol{E}_{u}^{k} $ to $l$. 
In cases where $\boldsymbol{E}_{u}^{k}$ exceeds the maximum length $l$, only the most recent $l$ behaviors are retained; otherwise, zero padding extends it to length $l$.
Positional encoding is then added:
\begin{equation}
    \boldsymbol{E}_{u}^k = \mathrm{PadOrTruncate}(\boldsymbol{E}_{u}^k; l) + \boldsymbol{P}^k,
\end{equation}
where $\boldsymbol{P}^k \in \mathbb{R}^{l\times d}$ denotes the positional embedding.

\subsubsection{Granularity-Aware Self-Attention} 
In order to enhance the model's ability to perceive the granularity of user behavior and capture the behavioral correlations within a single source, following~\cite{bert4rec}, we design a granularity-aware self-attention module~(GASA). GASA employs a multi-head self-attention to identify the correlations between behaviors of the same granularity. 
For $\boldsymbol{E}_{u}^{i}$, 
\begingroup
\small
\begin{equation}
\label{eq:self}
    \begin{split}
    \boldsymbol{Q}_i^h &= \boldsymbol{E}_{u}^{i} \cdot \boldsymbol{W}_i^{Qh}, \boldsymbol{K}_i^h = \boldsymbol{E}_{u}^{i} \cdot \boldsymbol{W}_i^{Kh}, \boldsymbol{V}_i^h = \boldsymbol{E}_{u}^{i} \cdot \boldsymbol{W}_i^{Vh} ,\\
    \mathrm{Attn}_i^h &= \mathrm{softmax}(\frac{\boldsymbol{Q}_i^h \cdot {(\boldsymbol{K}_i^h)}^\top}{\sqrt{d_k}}) \cdot \boldsymbol{V}_i^h, h \in [1,\cdots,H],\\
    \boldsymbol{g}_u^i &= \mathrm{Mean\_pooling}(\mathrm{Concat}(\mathrm{Attn}_i^1, ..., \mathrm{Attn}_i^H) \cdot \boldsymbol{W}_i^O)
    \end{split}
\end{equation}
\endgroup
where $\boldsymbol{W}_i^{Qh}, \boldsymbol{W}_i^{Kh}, \boldsymbol{W}_i^{Vh} \in \mathbb{R}^{d \times d_k}, \boldsymbol{W}_i^O  \in \mathbb{R}^{d \times d}$ are learnable weight matrices, $H$ is the number of attention heads, and $d_k = d/H$. 

\subsubsection{Cross-Granularity Fusion Attention}
Since the user's specific intentions may span multiple sources, we further design a cross-granularity fusion attention module~(CGFA) to capture the cross-source behavioral correlations. CGFA employs multi-head cross-attention to integrate the macro-level behavior patterns, as reflected in coarse-grained behaviors, with the micro-level insights derived from fine-grained behaviors. 
For $\boldsymbol{E}_{u}^{1}$, we compute

\begingroup
\small
\begin{equation}
    \begin{split}
    \boldsymbol{Q}_1^h &= \boldsymbol{E}_{u}^{1} \cdot \boldsymbol{W}_1^{Qh},  \boldsymbol{K}_{2,3}^h = \boldsymbol{E}_{u}^{2,3} \cdot \boldsymbol{W}_{2,3}^{Kh},  \boldsymbol{V}_{2,3}^h = \boldsymbol{E}_{u}^{2,3} \cdot \boldsymbol{W}_{2,3}^{Vh} ,\\
    \mathrm{Attn}_1^h &= \mathrm{softmax}(\frac{\boldsymbol{Q}_1^h \cdot {(\boldsymbol{K}_{2,3}^h)}^T}{\sqrt{d_k}}) \cdot \boldsymbol{V}_{2,3}^h, h \in [1,\cdots,H],\\
    \boldsymbol{c}_u^1 &= \mathrm{Mean\_pooling}(\mathrm{Concat}(\mathrm{Attn}_1^1, ..., \mathrm{Attn}_1^H) \cdot \boldsymbol{W}_1^O)
    \end{split}
\end{equation}
\endgroup
where  $\boldsymbol{W}_1^{Qh}, \boldsymbol{W}_{2,3}^{Kh}, \boldsymbol{W}_{2,3}^{Vh} \in \mathbb{R}^{d \times d_k}, \boldsymbol{W}_1^O  \in \mathbb{R}^{d \times d}$ are learnable weight matrices, $\boldsymbol{E}_{u}^{2,3}$ is the concatenation of $\boldsymbol{E}_{u}^2$ and $\boldsymbol{E}_{u}^3$, $H$ is the number of attention heads, and $d_k = d/H$. We can get $\boldsymbol{c}_u^2$ and $\boldsymbol{c}_u^3$ in a similar way as $\boldsymbol{c}_u^1$.

\subsubsection{User Representation}
Finally, we concatenate the outputs of GASA and CGFA to generate user representations with both source-specific and cross-source association knowledge.
For pre-training, to learn source-specific behavior patterns and avoid the negative influence of heterogeneous behaviors from other sources, we independently use the representations from the three sources to complete the pre-training task.
For downstream tasks, user representations are expected to encapsulate as much knowledge as possible, ensuring that downstream models can flexibly exploit rich local dependencies and global cross-source correlations.
Therefore, we concatenate all representations and feed the resulting vector into the downstream model. The representations of user $u \in \mathcal{U}$ are:
\begin{equation}
\label{eq:user_repr}
\begin{aligned}
   &\text{Pre-training}:~~~\boldsymbol{u}^k = \mathrm{Concat}(\boldsymbol{c}_u^k,\boldsymbol{g}_u^k),~k=1,2,3 \\
   &\text{Downstream}:~~~\boldsymbol{u} = \mathrm{Concat}(\boldsymbol{c}_u^1, \boldsymbol{c}_u^2, \boldsymbol{c}_u^3, \boldsymbol{g}_u^1, \boldsymbol{g}_u^2, \boldsymbol{g}_u^3)
\end{aligned}
\end{equation}

\subsection{Cross-Source Cascaded Prediction}
\label{subsec:prediction}

In our method, we propose a cross-source cascaded prediction framework that follows a hierarchical decision structure. Unlike parallel multi-task learning, the cascaded design explicitly conditions fine-grained predictions on high-level intent, which reduces the hypothesis space at each stage and mitigates label noise caused by cross-source ambiguity. 
Specifically, we decouple the standard next behavior prediction~(NBP) task~\cite{peterrec,uaf,ptum} into two sequential phases, an inter-source phase that predicts the source of the next behavior~(e.g., opening the AppStore), followed by an intra-source phase that refines the specific action within the selected source~(e.g., searching for a game). This cascaded design explicitly models high-level intent before fine-grained action, aligning with natural decision patterns. We detail each phase below.

(i)~\textbf{Inter-Source Phase.} 
This phase addresses the high-level task of \emph{behavior source prediction}~($T_1$), aiming to determine in which source~(e.g., GameApps, Appstore, Gamecenter) the user's next behavior will occur. Given a source-specific user embedding $\boldsymbol{u}^k$, we apply a shared source prediction tower $f_{\mathit{src}}(\cdot)$ to generate logits $\boldsymbol{r}_1^k$, followed by a softmax to obtain the predicted distribution $\hat{\boldsymbol{p}}_{\mathit{src}}^k$. To support subsequent tasks, an \textit{Info} module $f_{\mathit{info}}^1(\cdot)$ generates a context vector $\boldsymbol{z}_1^k$ summarizing the current task state as prior knowledge as follows. 
\begin{equation}
    \begin{split}
        \boldsymbol{r}_1^k &= f_{\mathit{src}}(\boldsymbol{u}^k), \\
        \hat{\boldsymbol{p}}_{\mathit{src}}^k &= \mathrm{softmax}(\boldsymbol{r}_1^k), \\
        \boldsymbol{z}_1^k &= f_{\mathit{info}}^1(\boldsymbol{r}_1^k),
    \end{split}
\end{equation}
During pre-training, the source label is derived from the held-out next behavior in the globally chronological multi-source log. Let $s_u \in \{1,2,3\}$ denote the source of this target behavior. We apply the same source supervision to each source-specific representation $\boldsymbol{u}^k$, encouraging each source encoder to capture cross-source transition patterns.
Here $f_{\mathit{src}}$(\textperiodcentered) denotes the source tower network and $f_{\mathit{info}}^1$(\textperiodcentered) refers to the \textit{Info} network; both are implemented as 2-layer MLPs with ReLU activation.

(ii)~\textbf{Intra-Source Phase.}
After modeling the behavior source, we next refine the intra-source behavior predictions.
Directly predicting the next action might be too difficult due to the large action space.
Therefore, we decompose the prediction of the user's specific action into three cascaded sub-tasks. Rather than passing hard predicted labels to subsequent tasks, the cascade propagates a differentiable task state $\boldsymbol{z}_i^k$: each sub-task conditions its prediction on the prior state $\boldsymbol{z}_{i-1}^k$ through an attention fusion step before softmax.
The sub-goals in this phase include \textit{App category prediction}~($T_2$), \textit{App ID prediction}~($T_3$), and \textit{Operation prediction}~($T_4$). We describe each sub-task below.

$T_2$ predicts the category of the App that the user will interact with next. 
$T_3$ predicts the specific App ID.
Once the App category is determined, the next step is to predict the exact App ID; conditioned on the category context from $T_2$, $T_3$ becomes more manageable.
$T_4$ predicts the specific operation type of the user's next behavior when operation labels are available and semantically meaningful. For Appstore and Gamecenter behaviors, both click and search events mainly reflect positive user intent in our setting; distinguishing them with $T_4$ therefore provides limited additional benefit for learning user interests, so this task is skipped to reduce computational complexity. In contrast, for third-party GameApps, install, uninstall, and launch operations can reflect different user attitudes, making $T_4$ useful for user interest modeling.

For each applicable sub-task $T_i$ $(i=2,3,4)$, a task-specific tower $f_T^i$ first maps the source-specific embedding $\boldsymbol{u}^k$ to features $\boldsymbol{r}_i^k$. The prior state $\boldsymbol{z}_{i-1}^k$---with $\boldsymbol{z}_1^k$ from $T_1$ when $i=2$---is fused with $\boldsymbol{r}_i^k$ via $f_{\mathit{attn}}^i$ to obtain the task context $\boldsymbol{q}_i^k$, from which $\hat{\boldsymbol{p}}_i^k$ is predicted. For $i \in \{2,3\}$, $f_{\mathit{info}}^i$ updates $\boldsymbol{z}_i^k$ for the next sub-task; when $i=4$, $\boldsymbol{z}_4^k$ is omitted because $T_4$ is the last intra-source sub-task on GameApps. Formally,
\begin{equation}
\label{eq:intra_cascade}
    \begin{split}
        \boldsymbol{r}_i^k &= f_T^i(\boldsymbol{u}^k), \\
        \boldsymbol{q}_i^k &= f_{\mathit{attn}}^i(\boldsymbol{r}_{i}^k, \boldsymbol{z}_{i-1}^k), \\
        \hat{\boldsymbol{p}}_i^k &= \mathrm{softmax}(\boldsymbol{W}_i \boldsymbol{q}_i^k + \boldsymbol{b}_i), \\
        \boldsymbol{z}_i^k &= f_{\mathit{info}}^i(\boldsymbol{q}_{i}^k), \quad i \in \{2,3\},
    \end{split}
\end{equation}
where $i$ indexes the applicable intra-source tasks among $\{2,3,4\}$, and $\boldsymbol{W}_i$, $\boldsymbol{b}_i$ are task-specific output parameters.
Here $f_T^i$(\textperiodcentered) represents the network of the $T_i$ tower and $f_{\mathit{info}}^i$(\textperiodcentered) denotes the $i$-th \textit{Info} module; $f_{\mathit{src}}$, $f_{\mathit{info}}^i$, and $f_T^i$ are all implemented as 2-layer MLPs with ReLU activation. $f_{\mathit{attn}}^i$(\textperiodcentered) represents the $i$-th \textit{Attention} module, implemented by self-attention.

\subsection{Model Training}
\label{subsec:train}

Let $\mathcal{A}_1$, $\mathcal{A}_2$, $\mathcal{A}_3$ represent the prediction task sets for third-party GameApps, Appstore, and Gamecenter in the intra-source stage, respectively. Specifically, $\mathcal{A}_1=\{T_2,T_3,T_4\}$ for third-party GameApps, while $\mathcal{A}_2=\mathcal{A}_3=\{T_2,T_3\}$ for Appstore and Gamecenter, where $T_4$ is omitted for the latter two sources as discussed above. Losses for the two stages can be expressed as follows:
\begin{equation}
    \mathcal{L}_{\mathit{INTER}} = -\sum_{k=1}^{3} \mathrm{log}(\hat{p}_{\mathit{src},s_u}^{k}),
\end{equation}
\begin{equation}
    \mathcal{L}_{\mathit{INTRA}} = -\sum_{k=1}^{3} \sum_{T \in \mathcal{A}_k} \sum_{i=0}^{N(T)-1}  (y^{k}_{T} = i) \mathrm{log}(\hat{p}_{T}^{k}),
\end{equation}
where $N(T)$ denotes the vocabulary size for sub-task $T$, i.e., the total number of candidate categories for that task; for example, when $T$ is App ID prediction~($T_3$), $N(T_3)$ is the number of unique candidate apps in the corresponding behavior source. $\hat{p}_{T}^{k}$ denotes the predicted probability distribution for task $T$ on source $k$.

The overall loss function of our pre-training tasks is:
\begin{equation}
    \mathcal{L} = \lambda \mathcal{L}_{\mathit{INTER}} + (1-\lambda) \mathcal{L}_{\mathit{INTRA}}, \label{eq:loss}
\end{equation}
where $\lambda$ is a hyper-parameter balancing the two stages.

To reduce computational complexity, we only set training and validation sets during the pre-training stage, and evaluate the effectiveness of user representations in downstream scenarios.

\section{Experiments}\label{sec:exp}
We conduct extensive experiments on large-scale real-world industrial datasets to evaluate the effectiveness of our proposed CM-PTM.
CM-PTM follows a distinct two-stage paradigm rather than end-to-end multi-task supervised learning.
In Stage~1, the model is trained via self-supervised mask-then-predict proxy tasks to learn universal user representations.
In Stage~2~(downstream adaptation), the pre-trained representations are frozen or fine-tuned and fed into a LightGBM~\cite{lightgbm} model together with user profile features for downstream mobile game recommendation. 

\subsection{Experimental Setting}
\subsubsection{Datasets} 

All datasets in our experiments are collected from OPPO, one of the largest smartphone manufacturers in the world, including user behaviors over the 180 days preceding the sampling date. The data is strictly anonymized and fully authorized by users under signed agreements, ensuring compliance with data privacy and security regulations. 
User behaviors consist of the following sources and types: `Launch', `Install' and `Uninstall' for third-party game apps; `Search' and `Click' for both App Store and Game Center.

\textbf{Pre-training Dataset.} 
We select a one-week sampling interval for pre-training from 2023/07/13 to 2023/07/19 to ensure data richness and diversity~(see Table~\ref{tab:pre-statistics} for details). 

\begin{table}[t]
  \centering
  \captionsetup{skip=4pt}
  \small
  \caption{Statistics of the pre-training dataset.}
  \label{tab:pre-statistics}
  {\setlength{\tabcolsep}{6.5pt}%
   \renewcommand{\arraystretch}{1}%
  \begin{tabular}{lc}
    \toprule
    \textbf{Statistics} & \textbf{Pre-training dataset} \\
    \midrule
    \# users & 1,947,683 \\
    \# apps & 83,341 \\
    \textit{avg}. \# gameapp operation behaviors & 144.93 \\
    \textit{avg}. \# appstore click behaviors & 62.21 \\
    \textit{avg}. \# appstore search behaviors & 22.60 \\
    \textit{avg}. \# gamecenter click behaviors & 3.84 \\
    \textit{avg}. \# gamecenter search behaviors & 1.12 \\
    \bottomrule
  \end{tabular}}
  \vspace{-0.8em}
\end{table}

\textbf{Downstream Task Datasets.}
We design downstream tasks on mobile game recommendation, and there is no temporal overlap between pre-training data and downstream task data. Although our experiments focus on the game domain, all evaluations are conducted on mobile game recommendation tasks defined below.

Specifically, we construct six distinct tasks~($G_1 \sim G_6$).  
We select popular blockbuster games for these tasks to ensure sufficient positive and negative interaction samples within the strict sampling windows, enabling robust and statistically significant offline evaluations.

Before describing the tasks, we first clarify several key concepts. \textbf{Old games} refer to games that were released at least six months ago. \textbf{New games} refer to games released within the past month. \textbf{Old users} are users who played the game during the past six months but not within the last two weeks. \textbf{New users} are users who never played the game before. \textbf{Attract} means that the user is labeled as positive if they played the target game within the specified evaluation window.

The three kinds of downstream scenarios are as follows:
\noindent
(i)~\textbf{Attract new users for old games.}
This scenario refers to predicting whether a new user will play an old game. We select three popular old games: \textit{Mini World} $(G_1)$, \textit{Eggy Party} $(G_2)$ and \textit{Sky: Children of Light} $(G_3)$. 

The training sets for $G_1 \sim G_3$ were sampled from 2023/07/20 to 2023/07/22. 
Positive samples are new users who played the game during the sampling period, while negative samples are those who had not played the game by the end of that period. 
The positive-to-negative ratio is $1:2$.

The test sets were sampled from 2023/07/23 to 2023/07/25. 
We randomly selected millions of new users and labeled them based on whether they played the game in the following week.

\noindent
(ii)~\textbf{Attract old users for old games.}
This scenario refers to predicting whether an old user will play an old game again. We select a widely popular game, \textit{Honor of Kings} $(G_4)$. 

The training set for $G_4$ was sampled from 2023/07/23 to 2023/07/25. 
Positive samples include old users who played the game during the sampling period, while negative samples are those who had not played the game by the end of that period. 
The positive-to-negative ratio is $1:2$ as well.

The test set was sampled from 2023/07/26 to 2023/07/28, consisting of millions of randomly selected users disjoint from the training and validation sets, labeled by whether they played the game during the following week.
\noindent
(iii)~\textbf{Attract new users for new games.}
This scenario refers to predicting whether a new user will play a new game. We select two popular games released in July 2023: \textit{Crystal of Atlan} $(G_5)$ and \textit{Journey to West: Burning Soul} $(G_6)$. 

The training sets for $G_5$ and $G_6$ were sampled from 2023/07/24 to 2023/07/25. The sampling rules are the same as those in~(i) and (ii).

\begin{table}[!t]
  \centering
  \small
  \caption{Statistics of the datasets for downstream tasks.}
  \label{tab:down-statistics}
  {\setlength{\tabcolsep}{6.5pt}%
  \begin{tabular}{crrc}
    \toprule
    & \# \textbf{Training Set} & \# \textbf{Test Set} & \% \textbf{Test Pos. Rate} \\
    \midrule
    $G_1$ & 236,609 & 4,995,000 & 0.018\% \\
    $G_2$ & 766,978 & 4,995,000 & 0.087\% \\
    $G_3$ & 72,923 & 4,995,000 & 0.003\% \\
    \midrule
    $G_4$ & 1,950,447 & 2,850,000 & 0.579\% \\
    \midrule
    $G_5$ & 675,033 & 4,985,000 & 0.219\% \\
    $G_6$ & 364,235 & 4,973,234 & 0.024\% \\
    \bottomrule
  \end{tabular}}
\end{table}
Table~\ref{tab:down-statistics} records the statistical information for the downstream datasets mentioned above.

\begin{table*}[t]
  \centering
  \caption{Overall performance of user representations generated by various pre-training methods on downstream tasks.}
  \label{tab:performance}
  {\setlength{\tabcolsep}{5pt}%
  \begin{tabular}{*{13}{c}}
    \toprule
    \multirow{2}*{\textbf{\makecell{Pre-training \\ Method}}} & \multicolumn{2}{c}{\textbf{$G_1$}} & \multicolumn{2}{c}{\textbf{$G_2$}} & \multicolumn{2}{c}{\textbf{$G_3$}} & \multicolumn{2}{c}{\textbf{$G_4$}} & \multicolumn{2}{c}{\textbf{$G_5$}} & \multicolumn{2}{c}{\textbf{$G_6$}} \\
    \cmidrule(lr){2-3}\cmidrule(lr){4-5}\cmidrule(lr){6-7}\cmidrule(lr){8-9}\cmidrule(lr){10-11}\cmidrule(lr){12-13}
    & \textbf{AUC} & \textbf{R@P}$_{0.95}$ & \textbf{AUC} & \textbf{R@P}$_{0.95}$ & \textbf{AUC} & \textbf{R@P}$_{0.95}$ & \textbf{AUC} & \textbf{R@P}$_{0.95}$ & \textbf{AUC} & \textbf{R@P}$_{0.95}$ & \textbf{AUC} & \textbf{R@P}$_{0.95}$ \\
    \midrule
    None & 0.7920 & 0.3164 & 0.7543 & 0.2382 & 0.8553 & 0.4268 & 0.8008 & 0.2569 & 0.8906 & 0.4698 & 0.9730 & 0.8546\\
     \midrule
    Bert4Rec & \underline{0.8370} & 0.4579 & 0.8109 & \underline{0.3836} & 0.8779 & 0.5786 & 0.8546 & 0.3485 & 0.8819 & 0.5054 & \textbf{0.9867} & 0.9034\\
    PeterRec & 0.8258 & \underline{0.4681} & 0.8113 & 0.3376 & 0.8697 & \underline{0.5792} & \underline{0.8590} & \underline{0.3607} & \underline{0.8948} & \underline{0.5103} & 0.9829 & \underline{0.9201}\\
    PTUM & 0.8235 & 0.4458 & \underline{0.8126} & 0.3415 & 0.8676 & 0.5244 & 0.8263 & 0.2969 & 0.8911 & 0.4729 & 0.9822 & 0.8710\\
     \midrule
    CLUE & 0.8198 & 0.4476 & 0.8036 & 0.3791 & 0.8689 & 0.5183 & 0.8347 & 0.3126 & 0.8916 & 0.4793 & 0.9803 & 0.8839\\
    CCL & 0.8215 & 0.4563 & 0.8097 & 0.3814 & 0.8721 & 0.5253 & 0.8406 & 0.3263 & 0.8921 & 0.4907 & 0.9835 & 0.8864\\
    AdaptSSR & 0.8247 & 0.3969 & 0.8017 & 0.3152 & \underline{0.8871} & 0.5548 & 0.8325 & 0.3236 & 0.8509 & 0.4252 & 0.9793 & 0.9190
    \\
     \midrule
     MAFN & 0.8366 & 0.4571 & 0.8064 & 0.3528 & 0.8679 & 0.5633 & 0.8367 & 0.3575 & 0.8859 & 0.4916 & 0.9856 & 0.9072 \\
     \midrule
    CM-PTM & \textbf{0.8542} & \textbf{0.4829} & \textbf{0.8411} & \textbf{0.4091} & \textbf{0.8976} & \textbf{0.5915} & \textbf{0.8724} & \textbf{0.3817} & \textbf{0.9056} & \textbf{0.5176} & \underline{0.9859} & \textbf{0.9288}\\
    \textit{Improvement} & +1.72\% & +1.48\% & +2.85\% & +2.55\% & +1.05\% & +1.23\% & +1.34\% & +2.10\% & +1.08\% & +0.73\% & -0.08\% & +0.87\% \\
    \bottomrule
  \end{tabular}}
\\[0.25em]
\noindent \footnotesize
\textit{\textbf{Bold} font indicates the best-performing method. \underline{Underline} indicates the second-best results. `None' indicates that only the profile features are input to LightGBM, with no user representation provided. Improvements over the best baselines are significant (paired t-test over 5 runs, p-value $<$ 0.05).}
\vspace{-0.8em}
\end{table*}

\subsubsection{Baselines}

We compare CM-PTM with representative generative pre-training methods and discriminative pre-training methods.
(a) Generative User Representation Pre-training Methods. \textbf{PeterRec} \cite{peterrec} applies masked behavior prediction, and \textbf{PTUM} \cite{ptum} applies masked behavior prediction and next-$K$-behavior prediction to the behavior sequence. \textbf{BERT4Rec} \cite{bert4rec} uses bidirectional self-attention to model the behavior sequence. 
(b) Discriminative User Representation Pre-training Methods. \textbf{CLUE}~\cite{clue} uses implicitly augmented views for contrastive pre-training. \textbf{CCL}~\cite{ccl} proposes a mask-and-fill data augmentation strategy and uses contrastive learning to train the model. \textbf{AdaptSSR}~\cite{adaptssr} proposes an adaptive data augmentation self-supervised ranking method to model user representations.
(c) Integrated User Representation Pre-training Method. \textbf{MAFN}~\cite{zhang2024pre} adopts both generative and discriminative methods to learn users' long-term and short-term interests. All baselines are provided with the same raw behavior data. For methods that do not support multi-source inputs, behaviors are chronologically merged into a single sequence.

\subsubsection{Evaluation Metrics}

We choose two widely used metrics, \textbf{AUC} and \textbf{R@P$_N$}, to evaluate the model performance. 
The first metric, AUC, is the area under the ROC curve. 
The second metric, R@P$_N$, indicates the maximum recall achieved under the constraint that precision is at least $N$. 
In this work, we set $N = 0.95$. 

\subsubsection{Implementation Details}

Our pre-training model is implemented with PyTorch~\cite{pytorch} under a Linux environment. Adam~\cite{adam} is selected as the optimizer. Considering the results of Optuna optimization and previous works~\cite{adaptssr,ccl}, we set the embedding dimension $d$ to 64 and the user representation dimension $d_u$ to $6 \times 64$. The batch size, learning rate, maximum length of each sequence $l$, number of attention heads $H$, and $\lambda$ in Eq.~(\ref{eq:loss}) are set to $512$, $1e-4$, $256$, $2$, and $1/2$, respectively. During pre-training, the training and validation sets are split in a $9:1$ ratio.

\subsection{Overall Performance on Downstream Tasks}

Table~\ref{tab:performance} presents the performance comparison results.

First, for $G_1$ to $G_5$, CM-PTM significantly outperforms all baseline models, achieving an AUC improvement of 1.05\% to 2.85\%. This gain can be attributed to the synergy between the cross-source cascaded prediction framework and the hierarchical attention mechanism.

Second, generative pre-training methods generally outperform discriminative counterparts. This is because generative methods directly model the behavior generation process, encouraging the learning of temporal dependencies across heterogeneous or homogeneous sources. In contrast, discriminative methods typically rely on behavior perturbation, e.g., masking or replacing actions, which may disrupt cross-source correlations, such as substituting an AppStore search with a GameApp launch, or distort key behavioral intents, replacing installation with uninstallation as an example. These are conjectured to be possible reasons for degraded intent representation and prediction accuracy.

Finally, we observe that CM-PTM's AUC is suboptimal for $G_6$. However, its R@P$_{0.95}$ performance is still far ahead, indicating a better ability to identify potential positive samples. This ability is even more crucial in practical new game recommendation scenarios, as new games need to attract sufficient positive users to build momentum and reputation. 

On the whole, the results demonstrate the outstanding performance of CM-PTM in various game scenarios, attributed to its effective capability to capture user intentions across different sources.

\begin{table}[t]
  \centering
  \captionsetup{skip=4pt}
  \caption{The AUC and R@P$_{0.95}$ performance with different components removed. The best results are in bold.}
  \label{tab:ab}
  {\setlength{\tabcolsep}{2.5pt}%
   \renewcommand{\arraystretch}{1}%
  \begin{tabular}{cccccc}
    \toprule
    \multirow{2}{*}{\makecell{\textbf{Ablation}\\\textbf{Group}}} & \multirow{2}{*}{\textbf{Variants}} & \multicolumn{2}{c}{$G_1$} & \multicolumn{2}{c}{$G_5$} \\
    \cline{3-6}
    & & \textbf{AUC} & \textbf{R@P}$_{0.95}$ & \textbf{AUC} & \textbf{R@P}$_{0.95}$ \\
    \midrule
    None & CM-PTM & \textbf{0.8542} & \textbf{0.4829} & \textbf{0.8956} & \textbf{0.5176} \\
    \midrule
    \multirow{2}{*}{Data} & \textit{w/o} fine-grained & 0.8485 & 0.4723 & 0.8728 & 0.4404 \\
    & \textit{w/o} coarse-grained & 0.8278 & 0.4388 & 0.8717 & 0.4413 \\
    \midrule
    Sequence & \textit{w/o} multi-seq & 0.8366 & 0.4539 & 0.8804 & 0.4547 \\
    \midrule
    \multirow{5}{*}{Task} & \textit{w/o} $T_1$ & 0.8490 & 0.4775 & 0.8769 & 0.4440 \\
    & \textit{w/o} $T_2$ & 0.8515 & 0.4796 & 0.8934 & 0.4946 \\
    & \textit{w/o} $T_3$ & 0.8477 & 0.4664 & 0.8919 & 0.4861 \\
    & \textit{w/o} $T_4$ & 0.8534 & 0.4818 & 0.8825 & 0.4476 \\
    & \textit{w/o} cascaded & 0.8307 & 0.4435 & 0.8726 & 0.4393 \\
    \bottomrule
  \end{tabular}}
\end{table}
\vspace{-0.8em}

\subsection{Ablation Study}
\label{subsec:ablation}

Next, we conduct an ablation study to dissect the contributions of different components of CM-PTM, categorized into three groups: \textit{data ablation}, \textit{sequence ablation}, and \textit{task ablation}. Table~\ref{tab:ab} presents results on both $G_1$ and $G_5$.

(i)~\textbf{Data ablation}. To examine the importance of jointly modeling multi-granularity behaviors from multiple sources, we selectively input either fine-grained or coarse-grained behavior data while keeping all other settings fixed. As shown in Table~\ref{tab:ab}, using only fine-grained behaviors decreases AUC by 2.64 and 2.39 percentage points on $G_1$ and $G_5$, respectively, while using only coarse-grained behaviors decreases AUC by 0.57 and 2.28 percentage points. These results robustly demonstrate that combining behaviors of different granularities from multiple sources is essential for accurately capturing user representations in mobile device scenarios.

(ii)~\textbf{Sequence ablation}. To investigate the limitations of single-sequence modeling, we construct a unified behavior sequence by chronologically merging all available behaviors and employ self-attention as the backbone model. The AUC drops by 1.76\% on $G_1$ and 1.52\% on $G_5$, highlighting the inadequacy of conventional single-sequence approaches~\cite{clue,zhang2024pre,adaptssr} in capturing the complexity of multi-granularity, cross-source behaviors in mobile environments.

(iii)~\textbf{Task ablation}. 
To validate the effectiveness of our cascaded prediction framework, we remove each sub-task $T_i$ from the task set ($T_1 \sim T_4$, denoted \textit{w/o} $T_i$) or replace the cascaded design with parallel predictions (\textit{w/o cascaded}) while keeping all other settings unchanged. 
As reported in Table~\ref{tab:ab}, $T_1$ contributes substantially, improving AUC by 0.52\% and 1.87\% for $G_1$ and $G_5$, respectively. 
For old games, among the four tasks, the contribution of $T_1$ ranks second only to App ID prediction~($T_3$), whose importance in user behavior modeling is self-evident. 
For new games, among the four tasks, removing~$T_1$ leads to the largest performance decrease.
Moreover, each sub-task as well as their cascade relationships contribute positively to the overall performance. 
These results support the rationale of our approach to capture cross-source intent through the hierarchical decision structure, especially behavior source prediction.
The cascaded design consistently outperforms the parallel alternative, demonstrating that progressively conditioning predictions on prior stages effectively captures cross-source dependencies across heterogeneous behavior streams.

\subsection{Case Study}

\begin{figure}[t]
  \centering
  \includegraphics[width=0.85\linewidth]{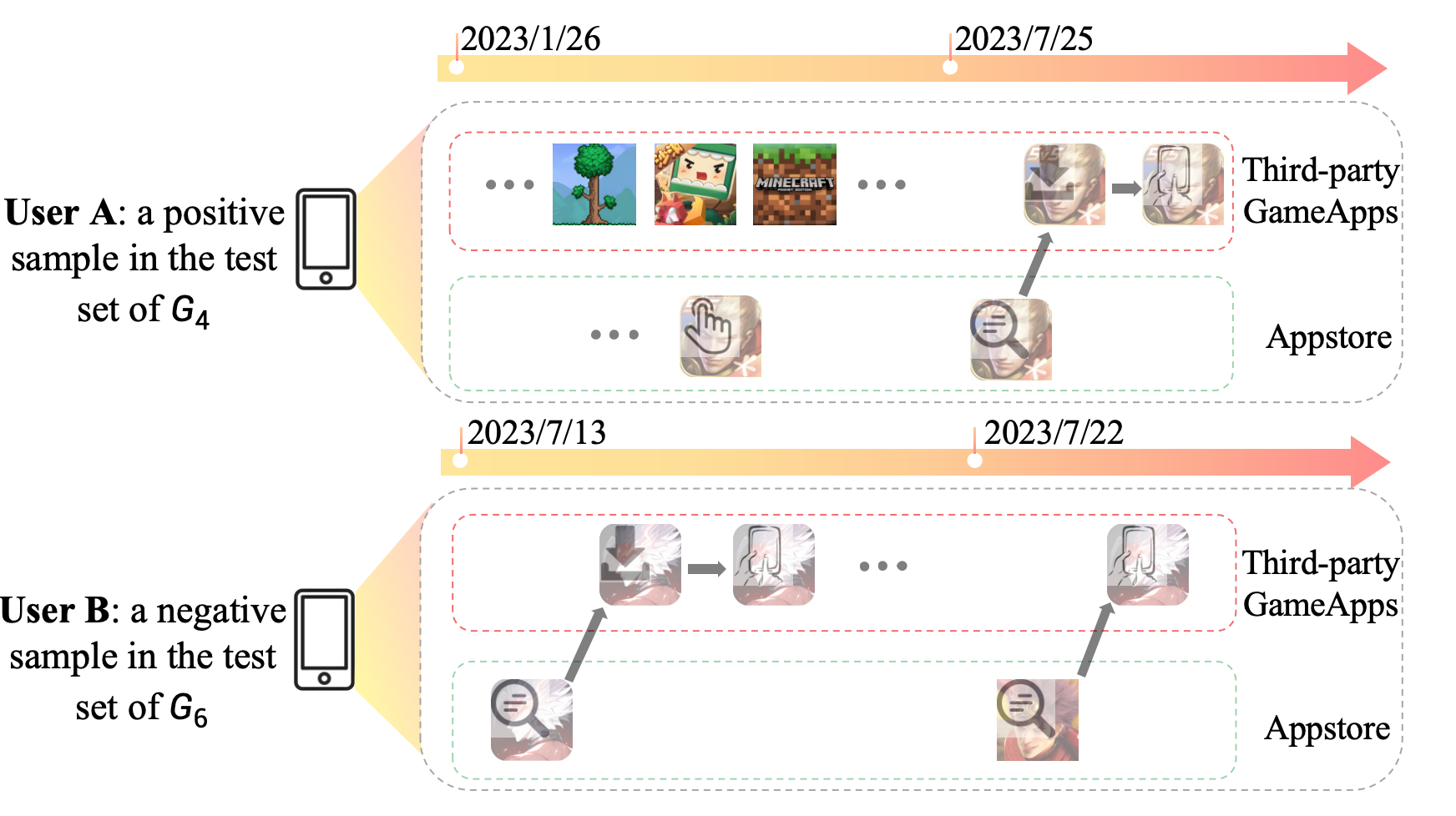}
  \caption{Two case study users and part of their behaviors.}
  \label{fig:case}
\end{figure}

We present two representative case studies in Figure~\ref{fig:case}. \textbf{User A} was correctly identified as a positive sample. From 2023/01/26 to 2023/07/24, this user had interacted with three sandbox games in third-party GameApps: \emph{Terraria}, \emph{Mini World}, and \emph{Minecraft}. Due to the strong preference for sandbox games, with only one click on \emph{Honor of Kings} in Appstore, our model successfully captured their interest in \emph{Honor of Kings}. This case illustrates the usefulness of cross-source intent modeling in mobile device environments.

\textbf{User B} was correctly identified as a negative sample. This user developed a strong interest in \emph{Crystal of Atlan}, searching for and downloading it on its release day~(2023/07/13). After the similar game \emph{Journey to West: Burning Soul} was released on 2023/07/22, this user searched for it without downloading it, returning to play \emph{Crystal of Atlan}. 
Through cross-source analysis, our model successfully identified the correlations between these behaviors, and then inferred that this user had limited interest in \emph{Journey to West: Burning Soul}. Notably, all baseline models produced false-positive predictions for this user, highlighting the superiority of our model in capturing cross-source user intent.

\subsection{Effects of Hyper-parameters}
\label{subsec:hyper}
\begin{figure*}[!t]
  \centering
  \subfloat[$G_2$]{%
    \includegraphics[width=0.20\textwidth]{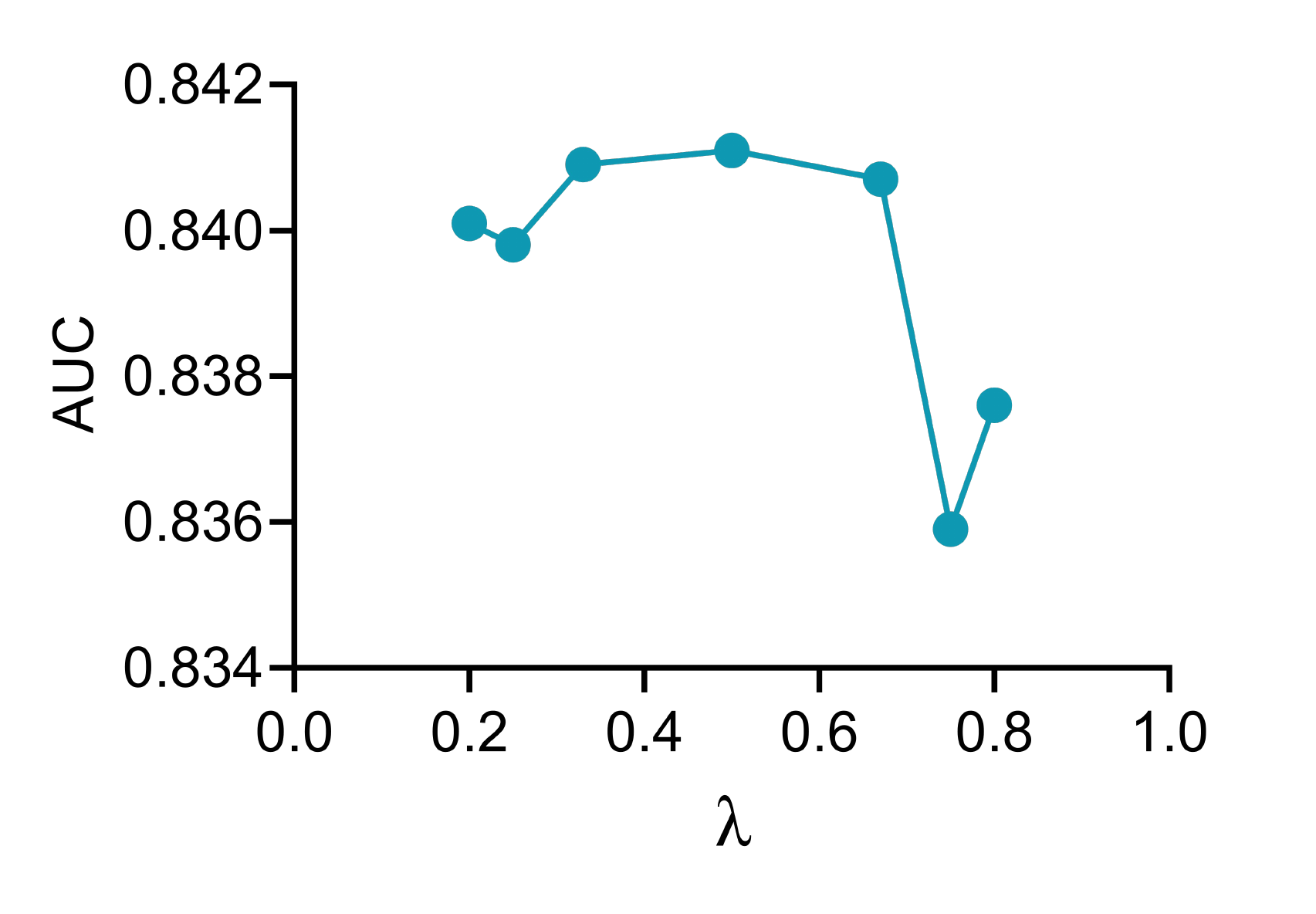}
  }%
  \hfill
  \subfloat[$G_3$]{%
    \includegraphics[width=0.20\textwidth]{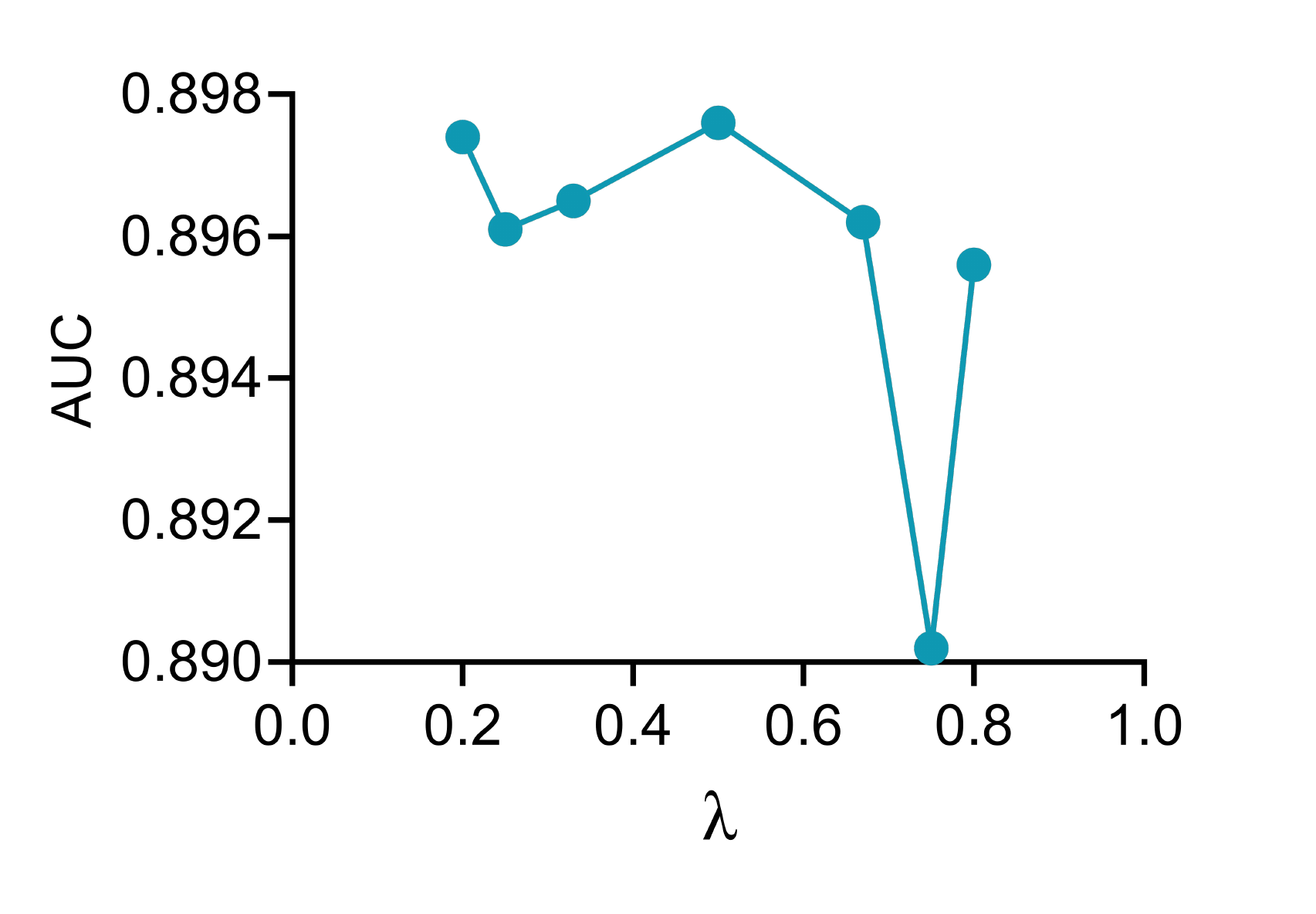}
  }%
  \hfill
  \subfloat[$G_4$]{%
    \includegraphics[width=0.20\textwidth]{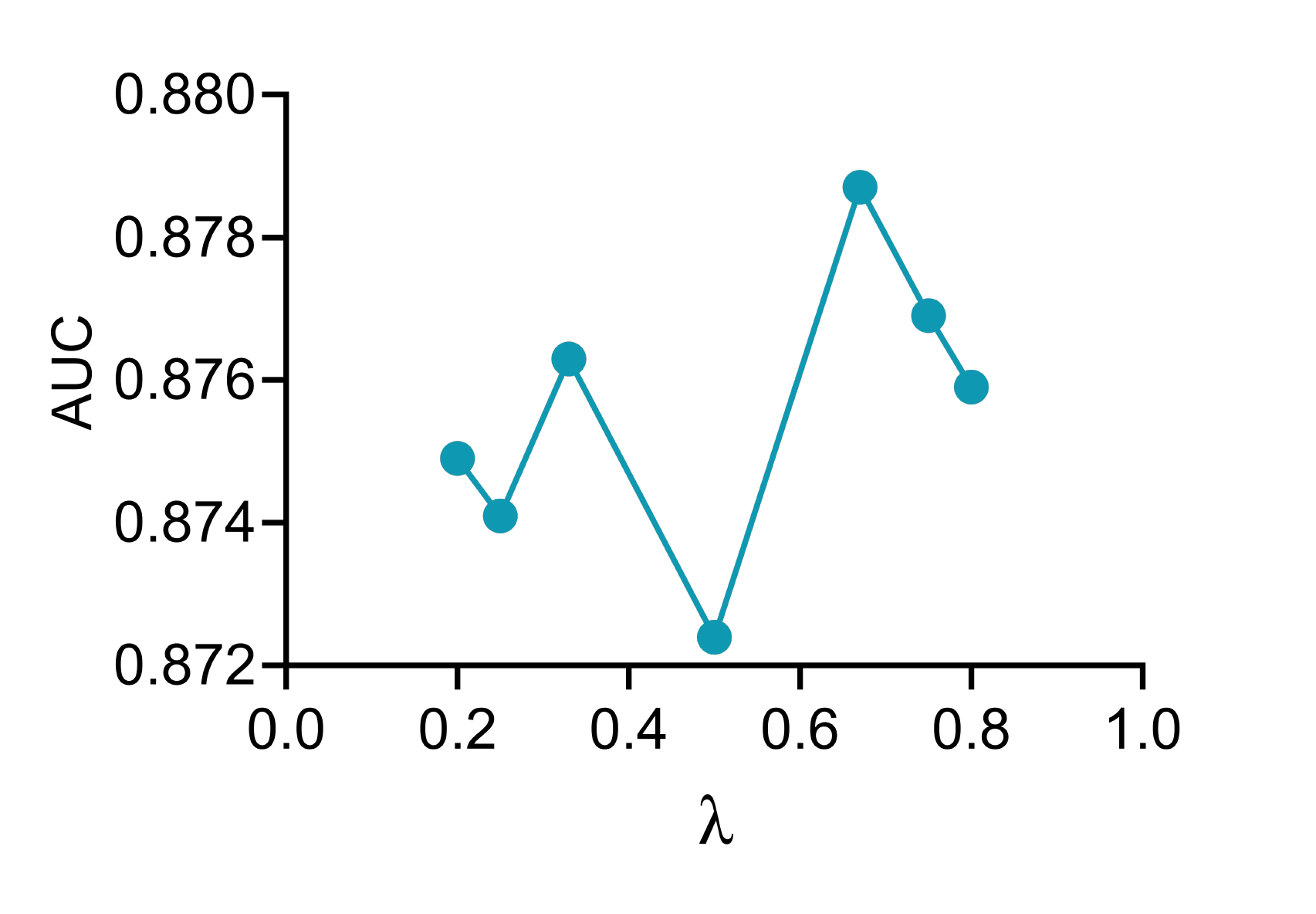}
  }%
  \hfill
  \subfloat[$G_5$]{%
    \includegraphics[width=0.20\textwidth]{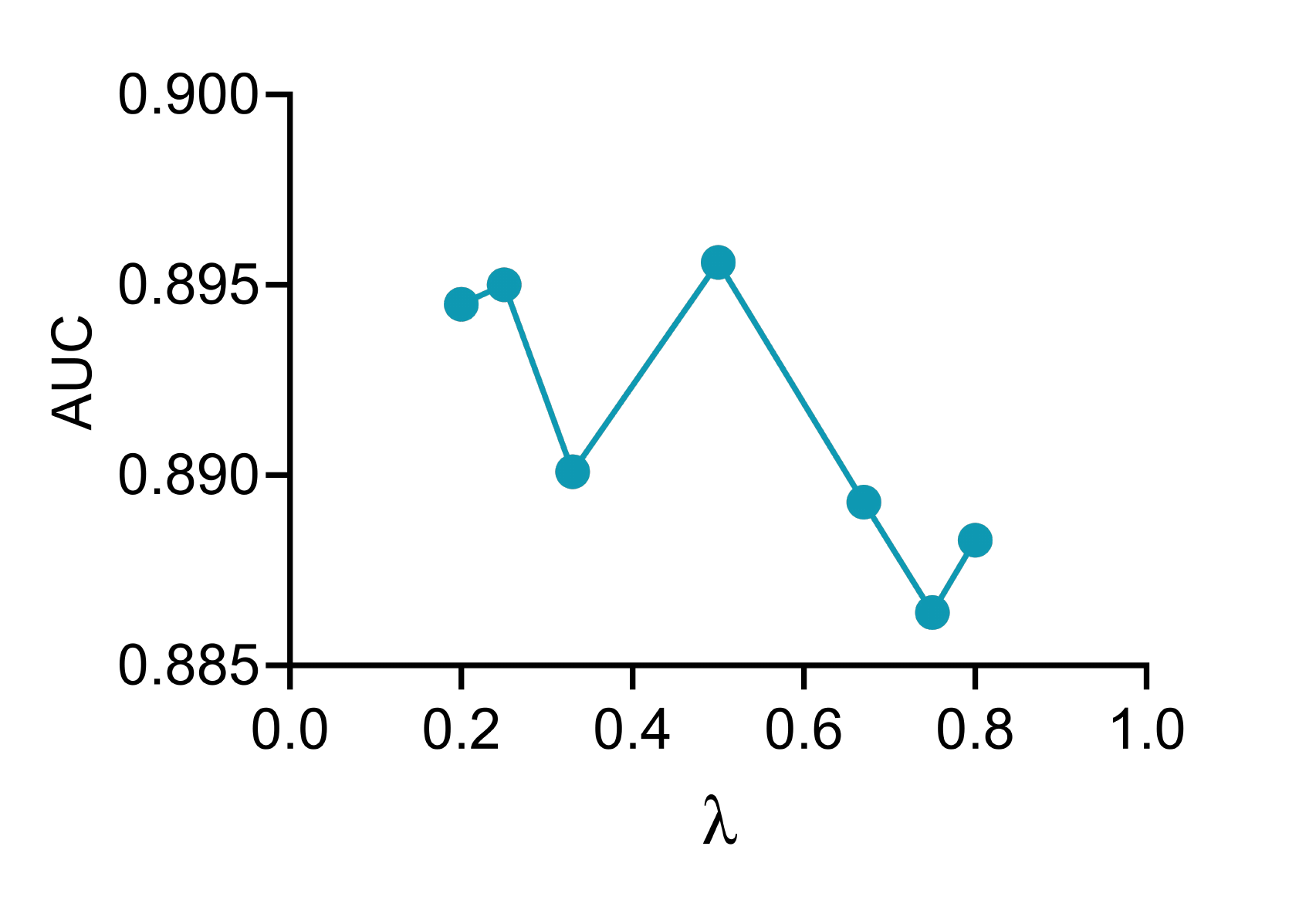}
  }%
  \caption{The impact of different values of $\lambda$.}
  \label{fig:lambda}
\end{figure*}

We set a hyper-parameter $\lambda$ in Section~\ref{subsec:train} to explore the relative importance of inter-source phase and intra-source phase. In this section, we select four downstream tasks to investigate the effects of different values of $\lambda$. Figure~\ref{fig:lambda} illustrates the results. 
For attracting new users, whether for an old or new game, setting $\lambda = 1/2$ performs best among the tested values, indicating that it is necessary to balance the optimization of the two phases. 
However, for old users, placing different emphasis on each phase may yield better outcomes. 
This is because old users have already formed their personal preferences: some users may be accustomed to downloading immediately after searching in the Appstore~(stable cross-source associations), while others may continuously browse related recommendations after searching and only download when they find something satisfying (uncertain cross-source associations).
Therefore, the model needs to prioritize the two phases to adapt to their established preferences.

\subsection{Online A/B Test}
\label{subsec:online}

\begin{table}[t]
  \centering
  \captionsetup{skip=2pt}
  \caption{Online A/B testing results.}
  \label{tab:online}
  {\setlength{\tabcolsep}{12pt}%
   \renewcommand{\arraystretch}{1}%
  \begin{tabular}{ccc}
    \toprule
    \textbf{Method} & \textbf{AUC} & \textbf{R@P}$_{0.95}$ \\
    \midrule
    DeepFM & 0.8064 & 0.3549 \\
    DeepFM + CM-PTM & 0.8192 & 0.3656 \\[0.2em]
    \hdashline
    \textit{Improvement} & \textbf{+1.28\%} & \textbf{+1.07\%} \\
    \bottomrule
  \end{tabular}}
\end{table}
\vspace{-0.8em} 

We also conduct an online A/B test on OPPO's mobile game recommendation system, and the results are shown in Table~\ref{tab:online}. Before deployment, CM-PTM was retrained on fresh, recent streaming behavioral data to reflect the latest user activity patterns. We randomly sampled 2.89 million active users over a two-week period~(ranging from 2025/01/05 to 2025/01/18) and split test traffic evenly by user ID between the two models to compare our CM-PTM against the baseline model DeepFM~\cite{deepfm}.  
Compared to DeepFM, our pre-training method improves AUC and R@P$_{0.95}$ by 1.28\% and 1.07\%, respectively, which validates the practicality of our method in real industrial scenarios on OPPO devices.

\subsection{Efficiency Comparison}

We finally compare the training time of CM-PTM with other baselines and analyze its online inference efficiency.

\begin{table}[t]
  \centering
  \captionsetup{skip=3pt}
  \caption{Offline training time for all compared methods.}
  \label{tab:time}
  {\setlength{\tabcolsep}{4.5pt}%
   \renewcommand{\arraystretch}{1}%
  \begin{tabular}{ccccc}
    \toprule
    \textbf{Method} & CM-PTM & Bert4Rec & PeterRec & PTUM \\
    \midrule
    \textbf{Training time}~(h) & 5.74 & 3.86 & 4.32 & 3.31 \\
    \midrule
    \textbf{Method} & CLUE & CCL & AdaptSSR & MAFN \\
    \midrule
    \textbf{Training time}~(h) & 3.02 & 2.87 & 3.24 & 6.53 \\
    \bottomrule
  \end{tabular}}
\end{table}
For offline training, all methods are pre-trained on two NVIDIA A100 GPUs. As shown in Table~\ref{tab:time}, CM-PTM requires more training time than most generative baselines but remains below MAFN and within an acceptable industrial training budget. This is because we adopt several strategies to reduce computational complexity: (1) no separate test set is constructed during pre-training; (2) the operation prediction task~($T_4$) is skipped for Appstore and Gamecenter behaviors; and (3) although each source-specific embedding $\boldsymbol{u}^k~(k=1,2,3)$ is pre-trained independently, its dimensionality is set to only 1/3 of $d_u$, introducing little additional computational burden. These strategies allow CM-PTM to maintain competitive training efficiency while modeling multi-source, multi-granular behaviors.

For online inference, CM-PTM augments the downstream representations by only $d_u$ dimensions, which is negligible compared to the existing thousand-dimensional features, yet consistently delivers substantial performance gains. This demonstrates that the model introduces minimal overhead in real-time inference.

In summary, CM-PTM incurs higher training cost than most generative baselines yet remains within an acceptable industrial budget, while introducing minimal inference overhead and providing significant improvements in downstream effectiveness, highlighting its practicality for large-scale industrial deployment.

\section{Conclusion}\label{sec:conclude}

In this paper, we propose CM-PTM, a novel pre-training framework for learning device-level user representations for mobile game recommendation from multi-source, multi-granular behavioral data on mobile devices. Inspired by a hierarchical decision structure for progressive intent refinement, CM-PTM decomposes behavior modeling into cascaded proxy tasks that capture both inter-source dependencies and fine-grained intra-source patterns in a progressive manner. Extensive experiments on large-scale industrial datasets from OPPO for mobile game recommendation demonstrate that CM-PTM consistently outperforms strong baselines in offline evaluations and online A/B tests, highlighting its effectiveness in capturing cross-source user intent and its practical applicability in real-world OPPO recommendation scenarios.
While our current model relies on positional embeddings to capture relative sequential order within short-term windows, explicitly incorporating time-aware mechanisms~(e.g., temporal encoding of absolute time intervals) is a promising direction for future work.
Our work provides a foundation for future research in device-level user representation pre-training for mobile games.

\section*{Ethics Statement}
This study was conducted in collaboration with OPPO Research Institute. The study was approved by OPPO's internal data governance board. Data were anonymized before modeling, and no raw textual search queries or content-level features were used. All data were collected with explicit user consent during service initialization. The dataset is restricted to three behavior sources at the mobile-device level: third-party GameApps, where only coarse-grained external signals such as installation, uninstallation, and launching are utilized, excluding sensitive in-app content and fine-grained interaction logs; App Store and Game Center, where data usage is limited to search and click-through events within these system applications. We only model event types and anonymized identifiers and deliberately adopt a design based on aggregated high-level cross-source signals to minimize privacy risks.

\section*{Acknowledgments}

The research work is supported by the National Natural Science Foundation of China under Grant Nos. 62576333, 62406307, and U2436209, the Strategic Priority Research Program of the Chinese Academy of Sciences under Grant No. XDB0680201, the Beijing Natural Science Foundation (F251001), and the Innovation Funding of ICT, CAS under Grant No. E461060.

\bibliographystyle{IEEEtran}
\bibliography{reficdm}

\end{document}